\documentclass[
]{ceurart}

\usepackage{listings}
\usepackage{enumitem}
\usepackage{todonotes}

\usepackage{relsize,setspace}
\usepackage{scalerel}

\usepackage{amsfonts}
\usepackage[mathcal]{euscript}
\usepackage{cleveref}
\usepackage{textcomp}

\usepackage{listings}
\newcommand{\set}[1]{\{ #1 \}}

\newcommand{\SROIQ}{\mathcal{SROIQ}}
\newcommand{\standpointLabel}{\texttt{standpointLabel}}
\newcommand{\Stands}{\mathcal{S}}
\newcommand{\xml}[1]{%
    \text{\normalfont\ttfamily\detokenize{#1}}%
}

\newcommand{\SSROIQ}{\mathbb{S}_{[\mathcal{SROIQ}]}}

\newcommand{\straightquote}{\textup{\texttt{\textquotedbl}}}

\newcommand{\myparagraph}[1]{\vspace*{-2.6ex}\paragraph{#1}}

\newcommand{\ebnfeq}{\mathrel{:\mkern-1mu:=}}

\newcommand{\Dom}{\Delta}

\def\dland{\sqcap}
\def\dlor{\sqcup}
\def\dlsub{\sqsubseteq}

\newcommand{\Preds}{\mathcal{P}}
\newcommand{\Consts}{\mathcal{C}}

\newcommand{\StandExps}{\mathcal{E}_{\mathcal{S}}}

\newcommand{\pred}[1]{\small\tt #1}

\newcommand{\Precs}{\Pi}

\newcommand{\pr}{\pi}
\renewcommand{\models}{\vDash}

\newcommand{\Inter}{\mathcal{I}} 

\newcommand{\atleast}[1]{\mathord{\geqslant}#1\,}
\newcommand{\atmost}[1]{\mathord{\leqslant}#1\,}

\newcommand{\conc}[1]{#1}
\newcommand{\rol}[1]{#1}

\newcommand{\rolexpR}{\rol{r}}

\newcommand{\rolR}{\mathtt{r}}
\newcommand{\rolS}{\mathtt{s}}

\newcommand{\rolU}{\mathtt{u}}

\newcommand{\conA}{\mathtt{A}}

\newcommand{\conC}{\conc{C}}
\newcommand{\conD}{\conc{D}}

\newcommand{\SROIQbs}{\ensuremath{\mathcal{SROIQ}b_s}\xspace}
\newcommand{\sroiq}{\text{$\mathcal{SROIQ}$}\xspace}

\newcommand{\tad}{\hspace{1pt}}

\newcommand{\Self}{\ensuremath{\mathit{Self}}}

\def\standb#1{\mathop{\Box}\nolimits_{\spform{#1}}}
\def\standd#1{\mathop{\Diamond}\nolimits_{\spform{#1}}}

\def\standbx#1{\mathop{\Box}\nolimits_{\scaleobj{0.8}{\spform{#1}}}}
\def\standdx#1{\mathop{\Diamond}\nolimits_{\scaleobj{0.8}{\spform{#1}}}}

\def\standbe{\standb{e}}

\def\standde{\standd{e}}

\def\Stands{\mathcal{S}}
\def\sts{\spform{s}} 
\def\Preds{\mathcal{P}}

\def\Consts{\mathcal{C}}

\def\E{\mathcal{E}}
\def\StandExps{\E_{\Stands}}
\def\ste{\spform{e}} 

\def\Dom{\Delta}

\newcommand{\spform}[1]{\mathsf{#1}}

\newcommand{\mycoloneq}{\mathbin{\,{:}\mkern-4mu{:}\mkern-1mu{=}\,}}

\newtheorem{example}{Example}

\newcommand{\FormatArithmeticComplexityClass}[1]{\ensuremath{\textsc{#1}}\xspace}
\newcommand{\NP}{\FormatArithmeticComplexityClass{NP}}

\newcommand{\mathcom}[3]{ \newcommand{#1}[#2]{\mbox{$#3$}}}
\mathcom{\imp}{0}{\ \rightarrow\ }            
\mathcom{\rimp}{0}{\ \leftarrow\ }            

\mathcom{\con}{0}{\ \wedge\ }                 
\mathcom{\dis}{0}{\ \vee\ }                   
\mathcom{\n}{0}{\neg}                     
\mathcom{\dimp}{0}{\ \leftrightarrow\ }       
\mathcom{\corresponds}{0}{\ \Lleftarrow\! \! \Rrightarrow\ }
\mathcom{\A}{0}{\forall}                  

\def\Box{\mathop\square}
\def\Diamond{\mathop\lozenge}

\def\star{\hbox{$*$}\xspace}

\def\pr{\pi}

\def\Precs{\Pi}

\def\standb#1{\Box\nolimits_{\spform{#1}}}
\def\standd#1{\Diamond\nolimits_{\spform{#1}}}

\def\standbx#1{\Box\nolimits_{\scaleobj{0.8}{\spform{#1}}}}
\def\standdx#1{\Diamond\nolimits_{\scaleobj{0.8}{\spform{#1}}}}

\def\standbe{\standb{e}}

\def\standde{\standd{e}}

\def\allstandb{\standb{*}}

\xspace
\def\f\xspacestandtopre{\hbox{$\sigma\,$}\xspace}
\def\fpretov\xspacealue{\hbox{$\delta\,$}\xspace}

\def\ModSat#1||-#2{#1\models #2}
\def\NotModSat#1||-#2{#1\nvDash #2}

\begin{document}

\copyrightyear{2023}
\copyrightclause{Copyright for this paper by its authors.
  Use permitted under Creative Commons License Attribution 4.0
  International (CC BY 4.0).}

\conference{}

\title{Automated reasoning support for Stand\-point-OWL~2}


\author[1]{Florian Emmrich}[%
email=florian.emmrich1@tu-dresden.de,
]
\cormark[1]
\address[1]{Technische Universität Dresden, Germany}

\author[1]{Lucía {Gómez Álvarez}}

\author[1]{Hannes Strass}



\cortext[1]{Corresponding author.}

\begin{abstract}
  We present a tool for modelling and reasoning with knowledge from various diverse (and possibly conflicting) viewpoints.
  The theoretical underpinnings are provided by enhancing base logics by \emph{standpoints} according to a recently introduced formalism that we also recall.
  The tool works by translating the standpoint-enhanced version of the description logic $\sroiq$ to its plain (i.e.\ classical) version.
  Existing reasoners can then be directly used to provide automated support for reasoning about diverse standpoints.
\end{abstract}

\begin{keywords}
  Standpoint Logic \sep
  OWL 2 DL \sep
  Reasoning
\end{keywords}

\maketitle

\section{Introduction}
The Semantic Web has democratised the production of knowledge sources by providing a set of standards for the specification of vocabularies, rules, and data stores.
The standard for authoring ontologies and knowledge bases is the \emph{Web Ontology Language} OWL~2~\cite{owl2overview:2012}, a language based on description logic (DL) \cite{baader_horrocks_lutz_sattler_2017}. 
Beyond the publication of independently developed sources, a fundamental goal of the Semantic Web is to support the integration and combination of the knowledge embedded within them.
However, the interoperability between ontologies is often hindered by semantic heterogeneity, differences in perspectives and other contextual factors. 

A recent proposal aiming to address these challenges is \emph{Standpoint Logic} (SL) \cite{alvarez:2021}, a framework for multi-perspective reasoning. SL is a multi-modal logic conceived to support the coexistence of multiple standpoints and the establishment of alignments between them. The language supports expressions of the form  $\standbx{s}[\phi]$ and $\standdx{s}[\phi]$, which express information relative to the \emph{standpoint} $\mathsf{s}$ and read, respectively: ``according to $\mathsf{s}$, it is \emph{unequivocal/conceivable} that $\phi$''.
In the semantics, standpoints are represented by sets of \emph{precisifications},\footnote{Precisifications are analogous to the \emph{worlds} of modal-logic frameworks with possible-worlds semantics.} such that $\standbx{s}[\phi]$ and $\standdx{s}[\phi]$ hold if $\phi$ is true in all/some of the precisifications in $\mathsf{s}$.
For the sake of illustration, let us revisit a condensed version of the example provided by~\citeauthor{alvarez:2022}~\cite{alvarez:2022}.

\begin{example}\label{example:fol}
	A range of conceptualisations for the notion of \emph{forest} have been specified for different purposes, giving rise to diverging or even contradictory statements regarding forest distributions.
	Consider a knowledge integration scenario involving two sources adopting a \emph{land cover} $(\mathsf{LC})$ and a \emph{land use} $(\mathsf{LU})$ perspective on forestry.
	$\mathsf{LC}$ characterises a \emph{forest} as a ``forest ecosystem'' with a minimum area \emph{(\ref{formula:defForestlandCover_sroiq})} where a \emph{forest ecosystem} is specified as an ecosystem with a certain ratio of tree canopy cover \emph{(\ref{formula:defForestEcosystem_sroiq})}.
	$\mathsf{LU}$ defines a forest with regard to the purpose for which an area of land is put to use by humans, i.e.\ a forest is a maximally connected area with ``forest use'' \emph{(\ref{formula:defForestlandUse_sroiq})}.\footnote{``Forest use'' areas may qualify for logging concessions and be classified into, e.g.\ agricultural or recreational use.}
	Sources $\mathsf{LC}$ and $\mathsf{LU}$ agree that forests subsume broadleaf, needleleaf and tropical forests \emph{(\ref{formula:forestSubclasses})}, and they both adhere to the Basic Formal Ontology \emph{($\mathsf{BFO}$) \cite{arp2015building}}, an upper-level ontology that formalises general terms, stipulating for instance that \emph{land} and \emph{ecosystem} are disjoint categories {\rm(\ref{formula:disjointLandEco})}.

Using standard DL notation and providing ``perspective annotations'' by means of correspondingly labelled multi-modal logic box operators, the example can be formalised in a stand\-point-enhanced description logic as follows:

\begin{enumerate}[series=SroiqForestry,label={\rm (F\arabic*)},ref={\rm F\arabic*},leftmargin=2em, itemsep=-1pt]\small
		\item $\standbx{LC}[\pred{Forest} \equiv \pred{ForestEcosystem}\dland \exists \pred{hasLand}. \pred{Area}_{\geq 0.5\mathrm{ha}}]$\label{formula:defForestlandCover_sroiq}
		\item $\standbx{LC}[\pred{ForestEcosystem}\equiv  \pred{Ecosystem} \dland \pred{TreeCanopy}_{\geq 20\%}]$\label{formula:defForestEcosystem_sroiq}
		\item $\standbx{LU}[\pred{Forest}\equiv\pred{ForestlandUse}\dland \pred{MCON}] \con \allstandb[ \pred{ForestlandUse}\sqsubseteq \pred{Land}]$\label{formula:defForestlandUse_sroiq}
		\item $\standbx{LC\cup LU}[(\pred{BroadleafForest}\dlor \pred{NeedleleafForest}\dlor \pred{TropicalForest}) \dlsub \pred{Forest}]$\label{formula:forestSubclasses}
		\item $(\mathsf{LC}\preceq\mathsf{BFO}) \con (\mathsf{LU}\preceq\mathsf{BFO}) \con \standbx{BFO}[\pred{Land}\dland\pred{Ecosystem}\dlsub\bot]$ \label{formula:disjointLandEco}
\end{enumerate}\vspace{-3ex}
\end{example}
Notice that \emph{ecosystem} and \emph{land} are disjoint categories according to the overarching $\mathsf{BFO}$ (\ref{formula:disjointLandEco}), yet forests are defined as ecosystems according to $\mathsf{LC}$ (\ref{formula:defForestlandCover_sroiq}) and as lands according to ${\mathsf{LU}}$ (\ref{formula:defForestlandUse_sroiq}). As discussed in \cite{alvarez:2021}, these kinds of disagreements result in well-reported challenges in the area of Ontology Integration  \cite{Euzenat2008OntologyAlignments,Otero-Cerdeira2015OntologyReview} and make ontology merging a non-trivial task.
Standpoint logic overcomes the usual tradeoffs by supporting standpoint-dependent knowledge specifications, which allows the statements {(\ref{formula:defForestlandCover_sroiq})--(\ref{formula:disjointLandEco})} to be jointly represented. 

In recent work, \citeauthor{alvarez:2022}~\cite{alvarez:2022} introduced \emph{First-Order} Standpoint Logic (FOSL) and showed favourable complexity results for its \emph{sentential} fragments, which disallow modal operators being applied to formulas with free variables. Specifically, adding sentential standpoints does not increase the complexity for fragments that are \emph{$\NP$-hard}, which is shown by means of a polytime equisatisfiable translation.  
These results apply to the sentential standpoint variants of the expressive $\SROIQ$ family of description logics, logical basis of OWL~2~DL \cite{owl2overview:2012}.
In a nutshell, given a knowledge base in sentential Standpoint-$\SROIQbs$\footnote{Notice that the published translation \cite{alvarez:2022} is for the mildly stronger $\SROIQbs$ instead of the more mainstream $\mathcal{SROIQ}$.}, the provided polytime translation outputs an equisatisfiable knowledge base in plain $\SROIQbs$. Beyond establishing tight complexity bounds, this presented us with a way to leverage existing highly optimised OWL reasoners to provide reasoning support for ontology languages extended by standpoint modelling. In this work, we adjust this translation to plain $\SROIQ$,\footnote{To the best of our knowledge current reasoners do not support $\SROIQbs$.} and we present an implementation thereof, which effectively constitutes the first tool supporting automated reasoning on Standpoint-OWL~2~DL in combination with existing off-the-shelf reasoners.

The paper is structured as follows. 
We first introduce the syntax and semantics of sentential Standpoint-$\SROIQ$ and describe briefly how $\SROIQ$ relates to OWL~2 (\Cref{section:StandpointDL}). We then explain how to encode sentential Standpoint-$\SROIQ$ axioms in an OWL~2~DL ontology, and how our implementation translates them in such a way that they can be processed by an OWL~2~DL reasoner (\Cref{section:StandpointOWL2}). We proceed to detail the usage of the command-line tool (\Cref{sec:tool-description}) and we conclude the paper with a discussion of the contributions and future work.

\section{Background}

We next introduce the theoretical background, starting with the “plain” (standpoint-free) description logic $\sroiq$, its standpoint-enhanced version, and the web ontology language OWL~2~DL.
\subsection{Standpoint Description Logic}\label{section:StandpointDL}
    \newcommand{\dlstruct}{\mathfrak{D}}
\newcommand{\standball}{\standb{*}}

Let $\Consts$, $\Preds_1$, and $\Preds_2$ be finite, mutually disjoint sets called \emph{individual names}, \emph{concept names} and \emph{role names}, respectively. $\Preds_2$ is subdivided into \emph{simple role names} $\Preds^\mathrm{s}_2$ and \emph{non-simple role names} $\Preds^\mathrm{ns}_2$, the latter containing the \emph{universal role} $\rolU$ and being strictly ordered by some strict order $\prec$.%
\footnote{In the original definition of $\mathcal{SROIQ}$, simplicity of roles and $\prec$ are not given a priori, but meant to be implicitly determined by the set of axioms. Our choice to fix them explicitly upfront simplifies the presentation without restricting expressivity.}
Then, the set $\mathcal{R}^\mathrm{s}$ of \emph{simple role expressions} is defined by
\mbox{\(
	\rolexpR_1,\rolexpR_2 \ebnfeq \rolS \mid \rolS^-
\)},
with $\rolS {\,\in\,} \Preds^\mathrm{s}_2$, while the set of (arbitrary) \emph{role expressions} is $\mathcal{R} {\,=\,} \mathcal{R}^\mathrm{s} {\,\cup\,} \Preds^\mathrm{ns}_2$\!. The order $\prec$ is then extended to $\mathcal{R}$ by making all elements of $\mathcal{R}^\mathrm{s}$ $\prec$-minimal.
The syntax of \emph{concept expressions} is given by
\mbox{\(
			\conC,\conD \ebnfeq \conA \mid \{a\} \mid \top \mid \bot \mid \neg\conC \mid \conC\sqcap\conD \mid \conC\sqcup\conD \mid \forall\rolexpR.\conC \mid \exists\rolexpR.\conC \mid \exists\rolexpR'\!.\mathit{Self} \mid \atmost{n}\rolexpR'\!.C \mid \atleast{n}\rolexpR'\!.C,
\)}
with \mbox{$\conA\in \Preds_1$}, \mbox{$a\in \Consts$}, \mbox{$\rolexpR \in  \mathcal{R}$}, \mbox{$\rolexpR' \in  \mathcal{R}^\mathrm{s}$}, and \mbox{$n\in \mathbb{N}$}.
%
The different types of $\mathcal{SROIQ}$ sentences (called \emph{axioms}) are given in Table~\ref{tab:axm}.\footnote{The original definition of \sroiq contained more axioms (role transitivity, (a)symmetry, (ir)reflexivity and disjointness; concept and role assertions, i.e., ABox axioms; (in)equality), but these are syntactic sugar in our setting.}
\newcommand{\tuplei}[1]{(#1)}
\begin{table}[t]
	\setlength{\tabcolsep}{-0.5em}
	\begin{tabular}{ll}
		{
			~\hspace{\stretch{1.5}}
			~\vspace{\stretch{1.5}}
			\!\!\!\!\scalebox{.91}{
    \begin{tabular}[t]{@{}l@{}r@{\ \ }r@{}}
						\hline                                                                                                                                                                                                                                                       \\[-2ex]
						Name                           & Syntax                                                                                & Semantics                                                                                                                           \\\hline\\[-2ex]
						general concept \hspace{-6ex}  & $C{\,\sqsubseteq\,} D$\!\!                                                            & $C^\mathcal{I}\subseteq D^\mathcal{I}$                                                                                              \\
						~inclusion (GCI) \hspace{-6ex} &                                                                                       &
						\\\hline\\[-2ex]
						role inclusion                 & $\rolexpR_1{\circ}\ldots{\circ}\tad\rolexpR_n{\,\sqsubseteq\,} \rolR$                 & $\rolexpR_1^\mathcal{I}{\circ}\ldots{\circ}\tad\rolexpR_n^\mathcal{I}{\,\subseteq\,} \rolR^\mathcal{I}$                             \\
						~axioms                        & $\rolexpR_1{\circ}\ldots{\circ}\tad\rolexpR_n{\circ}\tad\rolR{\,\sqsubseteq\,} \rolR$ & $\rolexpR_1^\mathcal{I}{\circ}\ldots{\circ}\tad\rolexpR_n^\mathcal{I}{\circ}\tad\rolR^\mathcal{I}{\,\subseteq\,} \rolR^\mathcal{I}$ \\
						~(RIAs)                        & $\rolR{\circ}\tad\rolexpR_1{\circ}\ldots{\circ}\tad\rolexpR_n{\,\sqsubseteq\,} \rolR$ & $\rolR^\mathcal{I}{\circ} \rolexpR_1^\mathcal{I}{\circ}\ldots{\circ}\tad\rolexpR_n^\mathcal{I}{\,\subseteq\,} \rolR^\mathcal{I}$    \\  			                                         & $\rolR{\circ}\tad\rolR{\,\sqsubseteq\,} \rolR$                                     & $\rolR^\mathcal{I}{\circ}\tad\rolR^\mathcal{I}{\,\subseteq\,} \rolR^\mathcal{I}$                                              \\\hline\\[-1ex]
						In RIAs, $\rolR \in \Preds^\mathrm{ns}_2$, while $\rolexpR_i \in \mathcal{R}$ and $\rolexpR_i \prec \rolR$ for all \hspace{-10cm}                                                                                                                            \\
						$i\in \{1,\ldots,n\}$.                                                                                                                                                                                                                                       \\
					\end{tabular}
			}\hspace{\stretch{1.5}}~
		}
		%
		%
		 & {
				~\hspace{\stretch{1.5}}
				~\vspace{\stretch{1.5}}

				\!\scalebox{.91}{
						\begin{tabular}[t]{@{}l@{\ \ \,}l@{\ \ \,}l@{}}
					\hline                                                                                                                                                              \\[-2ex]
					Name                & Syntax                            & Semantics                                                                                                 \\\hline
					inverse role        & $\rolS^-$                         & $\{\tuplei{\delta,\delta'}\in\Delta\times\Delta \mid \tuplei{\delta',\delta} \in \rolS^\Inter\}$          \\
					universal role      & $\rolU$                           & $\Delta^\Inter\times\Delta^\Inter$                                                                        \\
					\hline                                                                                                                                                              \\[-2ex]			                                                                                  nominal             & $\{a\}$                           & $\{a^{\Inter}\}$                                                                                          \\
					top                 & $\top$                            & $\Delta^\Inter $                                                                                          \\  
					bottom              & $\bot$                            & $\emptyset$                                                                                               \\  
					negation            & $\neg \conC$                      & $\Delta^\Inter \setminus \conC^{\Inter}$                                                                  \\  
					conjunction         & $\conC\sqcap \conD$               & $\conC^{\Inter}\cap \conD^{\Inter}$                                                                       \\  
					disjunction         & $\conC\sqcup \conD$               & $\conC^{\Inter}\cup \conD^{\Inter}$                                                                       \\  
					univ. restriction   & $\forall \rolexpR.\conC$          & $\{\delta \mid \forall y. \tuplei{\delta,\delta'} \in \rolexpR^{\Inter} \to \delta'\in \conC^{\Inter}\}$  \\  
					exist. restriction  & $\exists \rolexpR.\conC$          & $\{\delta \mid \exists y. \tuplei{\delta,\delta'}\in\rolexpR^{\Inter} \wedge \delta'\in \conC^{\Inter}\}$ \\  
					$\Self$ concept     & $\exists\rolexpR.\Self$           & $\{\delta \mid \tuplei{\delta,\delta}\in\rolexpR^{\Inter}\}$                                              \\
					qualified number    & $\atmost{n}\rolexpR.C$            & $\{\delta \mid \#\{\delta'\in \conC^{\Inter}\mid \tuplei{\delta,\delta'} \in \rolexpR^{\Inter}\}\le n\}$  \\
					\qquad restrictions & $\atleast{n}\rolexpR.C$           & $\{\delta \mid \#\{\delta'\in \conC^{\Inter}\mid \tuplei{\delta,\delta'} \in \rolexpR^{\Inter}\}\ge n\}$  \\
					\hline                                                                                                                                                              \\[-2ex]
				\end{tabular}
				}
			}

		\hspace{\stretch{1.5}}~
	\end{tabular}
\vspace{-1em}
	\caption{\small “Plain” $\mathcal{SROIQ}$ role, concept expressions, RIAs and TBox axioms. $C\equiv D$ abbreviates $C\sqsubseteq D$, $D\sqsubseteq C$.
	}\label{tab:axm}\label{tab:SROIQ}
	\vspace{-2em}
\end{table}

Similar to FOL, the semantics of $\mathcal{SROIQ}$ is defined via interpretations \mbox{$\mathcal{I}=(\Delta,\cdot^\mathcal{I})$} composed of a non-empty set $\Delta$ called the \emph{domain of~\,$\mathcal{I}$} and a function $\cdot^\mathcal{I}$ mapping individual names to elements of $\Delta$, concept names to subsets of $\Delta$, and role names to subsets of \mbox{$\Delta\times\Delta$}.
This is extended to role and concept expressions 
and used to define satisfaction of axioms (see Table~\ref{tab:axm}).

In Standpoint-\sroiq, “plain” \sroiq axioms may be preceded by a \emph{standpoint modality}, expressing a standpoint relative to which the axiom is stated to hold.
Within such modalities, standpoints may be either referred to by name (e.g.\ as in \ref{formula:defForestlandCover_sroiq}--\ref{formula:disjointLandEco}), or by expressions constructed from names inductively using set operators.
Formally, the set $\StandExps$ of \emph{standpoint expressions} is defined by
\mbox{\(
	\ste_1,\ste_2 \ebnfeq * \mid \sts \mid \ste_1\cup\ste_2 \mid \ste_1\cap\ste_2 \mid \ste_1\setminus\ste_2
\)}, where $s\in\Stands$ is a \emph{standpoint name}, and $*\in\Stands$ is a special name referring to the \emph{universal standpoint}, i.e.\ the standpoint comprising all precisifications.
A \emph{sharpening statement} \mbox{$\ste_1\preceq\ste_2$} expresses that $\ste_1$ pertains to a viewpoint that is at least as narrow as that of $\ste_2$ and is syntactic sugar for the axiom \mbox{$\standb{\ste_1\backslash\ste_2}[\top\sqsubseteq\bot]$}.
The set $\SSROIQ$ of \emph{sentential Stand\-point-$\mathcal{SROIQ}$ sentences} is now defined as the union
\mbox{\(
    \SSROIQ := \mathcal{B}_R \cup \mathcal{B}_T,
\)}
where $\mathcal{B}_R$ consists of all $\SROIQ$ RIAs and $\mathcal{B}_T$ is inductively defined:
\begin{itemize}[itemsep=-1pt]
    \item if $\phi$ is a $\SROIQ$ TBox axiom, then $\phi \in \mathcal{B}_T$,
    \item if $\phi,\psi \in \mathcal{B}_T$, then $\neg\phi, \phi \wedge \psi, \phi \vee \psi \in \mathcal{B}_T$,
    \item if $\phi \in \mathcal{B}_T$ and $\mathsf{e} \in \mathcal{E}_{\mathcal{S}}$, then $\Box_{\mathsf{e}}\phi, \Diamond_{\mathsf{e}}\phi \in \mathcal{B}_T$.
\end{itemize}
Any $\phi \in \SSROIQ$ can be transformed to an equivalent \mbox{$\psi \in \SSROIQ$} in \emph{normal form}, where negation only occurs directly before a $\SROIQ$ TBox axiom or a standpoint modality $\Box_{\mathsf{e}}$/$\Diamond_{\mathsf{e}}$, and no standpoint modality appears in the scope of another.

In the semantics of sentential Stand\-point-$\mathcal{SROIQ}$, standpoints are represented by sets of so-called \emph{precisifications} where each precisification corresponds to an ordinary $\mathcal{SROIQ}$ interpretation.
Formally, the semantics of (sentential) Standpoint-\sroiq knowledge bases $\mathcal{K}\subseteq\SSROIQ$ is given by \emph{description logic standpoint structures} $\dlstruct=\tuplei{\Dom, \Precs, \sigma, \gamma}$ where
$\Dom$ is a non-empty set, the interpretation \emph{domain},
$\Precs$ is a non-empty set of \emph{precisifications},
$\sigma$ maps each standpoint name $\sts\in\Stands$ to a subset of $\Precs$, and
$\gamma$ maps each $\pr\in\Precs$ to a “plain” \sroiq interpretation with domain $\Dom$.
The satisfaction relation for DL standpoint structures and elements of $\SSROIQ$ is then given by
\begin{itemize}[itemsep=-1pt]
    \item  \mbox{\(
\dlstruct,\pr\models\xi \text{ iff } \gamma(\pr)\models\xi
\)}
for $\SROIQ$ TBox axioms $\xi$, and
\item \mbox{\(
\dlstruct,\pr\models\standbe[\phi] \text{ iff } \dlstruct,\pr'\models\phi \text{ for each } \pr'\in\sigma(\ste)
\)} and
\item \mbox{\(
\dlstruct,\pr\models\standde[\phi] \text{ iff } \dlstruct,\pr'\models\phi \text{ for some } \pr'\in\sigma(\ste)
\)}
\end{itemize}
where
$\sigma$ is extended from standpoint names to standpoint expressions in the obvious way, and
the satisfaction relation for the Boolean connectives is as usual.
In a Standpoint-$\SROIQ$ knowledge base $\mathcal{K}\subseteq\SSROIQ$, we consider all formulas $\phi$ not preceded by a modality to be implicitly of the form $\standball[\phi]$.
For the full technical definitions we refer to the original paper~\cite{alvarez:2022}.

We finally note that G\'omez \'Alvarez et al.~\cite{alvarez:2022} have presented a sentential standpoint version of the description logic $\SROIQbs$, which extends $\SROIQ$ by \emph{safe Boolean role expressions}, i.e.\ role expressions of the form $r_1 \cup r_2$, $r_1 \cap r_2$ and $r_1 \setminus r_2$, denoting union, intersection and difference of roles, respectively. Since to our knowledge there is no reasoner which supports these safe Boolean role expressions, we have restricted the implementation to sentential Stand\-point-$\mathcal{SROIQ}$.
\subsection{OWL~2~DL}
    
The \emph{Web Ontology Language} \emph{OWL~2}~\cite{owl2overview:2012} is an expressive knowledge representation language and a W3C-recommended standard for modelling ontologies. There are two alternative ways of defining the semantics of OWL~2 ontologies: the \emph{RDF-Based Semantics}~\cite{owl2RDFSemantics:2012}, which assigns meaning to RDF graphs and thus only indirectly to ontology structures via the mapping to RDF graphs~\cite{owl2mapping:2012}, and the \emph{Direct Semantics}~\cite{owl2DirectSemantics:2012} which assigns meaning directly to ontology structures. The latter results in a semantics compatible with the model-theoretic semantics of $\SROIQ$. 

Moreover, to ensure that OWL~2 ontology structures can be translated into a $\SROIQ$ knowledge base, certain conditions have to be fulfilled, for instance transitive properties cannot be used in number restrictions. A complete list of restrictions can be found in the OWL~2 Structural Specification document~\cite[Section~3]{owl2structure:2012}. Ontologies that satisfy these conditions are called \emph{OWL~2~DL} ontologies.
Our focus is on OWL~2~DL since this compatibility with $\SROIQ$ ontologies allows us to implement the translation from sentential Stand\-point-$\mathcal{SROIQ}$ to standard $\SROIQ$ in OWL~2. 

\section{Standpoint-OWL~2~DL}\label{section:StandpointOWL2}

In order to support standpoint-based reasoning in the semantic web, one may either extend current standards such as OWL~2, or provide procedures to encode the standpoint operators within these languages. We take the latter approach following the lines of the work of Bobillo and Straccia~\cite{bobillo:2011}, who proposed a methodology to represent fuzzy ontologies in OWL~2 using annotation properties. These properties are broadly used to add comments or labels to entities and axioms of the ontology, as a way to provide supplementary information to the user.
In our case, we define the annotation property ``\standpointLabel'', which will be used to add standpoint operators to axioms and to create Boolean combinations of standpoint axioms. 


This section illustrates how to encode the sentential Stand\-point-$\SROIQ$ (Section \ref{section:StandpointDL}) constructs that are not available in OWL~2. Most importantly, we provide the syntax to encode Boolean combinations of standpoint axioms, i.e. the axioms in $\mathcal{B}_T$. While this is sufficient to encode standpoint ontologies, we also introduce syntax for the specification of sharpening statements, which are syntactic sugar in Stand\-point-$\SROIQ$, and also for labelling single standard OWL~2 subclass or equivalence axioms with a standpoint operator, which facilitates the enhancement of pre-existing ontologies with standpoints. 



\myparagraph{Boolean combinations}
Complex standpoint axioms can be added to an ontology by annotating the ontology itself by a \standpointLabel \ with a $\mathrm{BoolComb}$ value:\footnote{The elements in the XML syntax are not case-sensitive, but the name attribute is.}
\begin{align*}
    \mathrm{BoolComb}       \mycoloneq& \       \xml{<booleanCombination>} \mathrm{Formula} \xml{</booleanCombination>} \\
    \mathrm{Formula}        \mycoloneq& \       \mathrm{Axiom} \ | \
                                                \xml{<NOT>} \mathrm{Axiom} \xml{</NOT>} \ | \\
                            \phantom{=}& \      \xml{<AND>} \mathrm{Formula} \ \mathrm{Formula} \xml{</AND>} \ | \
                                                \xml{<OR>} \mathrm{Formula} \ \mathrm{Formula} \xml{</OR>} \\
    \mathrm{Axiom}          \mycoloneq& \       \mathrm{StdAxiom} \ | \
                                                \xml{<standpointAxiom name=}\straightquote\xml{§ax}\straightquote\xml{/>} \ | \\
                            \phantom{=}& \      \xml{<Box>} \mathrm{SPExpr} \ \mathrm{StdAxiom} \xml{</Box>} \ | \
                                                \xml{<Diamond>} \mathrm{SPExpr} \ \mathrm{StdAxiom} \xml{</Diamond>} \\
    \mathrm{StdAxiom}       \mycoloneq& \       \xml{<subClassOf> <LHS>} \mathrm{Class} \xml{</LHS> <RHS>} \mathrm{Class} \xml{</RHS> </subClassOf>} \ | \\
                            \phantom{=}&  \xml{<equivalentClasses> <LHS>} \mathrm{Class} \xml{</LHS>} \xml{<RHS>} \mathrm{Class} \xml{</RHS> </equivalentClasses>} \\
    \mathrm{SPExpr}         \mycoloneq& \       \xml{<Standpoint name=}\straightquote\xml{s}\straightquote\xml{/>} \ | \
                                                \xml{<INTERSECTION>} \mathrm{SPExpr} \xml{</INTERSECTION>} \ | \\
                            \phantom{=}& \      \xml{<UNION>} \mathrm{SPExpr} \xml{</UNION>} \ | \
                                                \xml{<MINUS>} \mathrm{SPExpr} \xml{</MINUS>}
\end{align*}
where $\xml{§ax}$ matches the regular expression $\xml{§[a-zA-Z]+[0-9]*}$, $\xml{s}$ either matches $\xml{[a-zA-Z]+[0-9]*}$ or is the universal standpoint $\xml{*}$, and $\mathrm{Class}$ is a class expression in OWL~2 Manchester syntax~\cite{horridge:2012}.
If a named standpoint axiom is mentioned, there has to be an annotated axiom with the same name attribute in the ontology, which then replaces the reference in the Boolean combination.

\begin{example}\label{example:annotation}
We can encode the axiom $($\ref{formula:defForestlandUse_sroiq}$)$ in \Cref{example:fol}
    by annotating the ontology with a \standpointLabel \ in the following way:
    \begin{small}\begin{lstlisting}
        $\xml{<standpointLabel>}$ $\xml{<booleanCombination>}$ $\xml{<AND>}$
            $\xml{<Box>}$ $\xml{<Standpoint name=}\straightquote\xml{LU}\straightquote\xml{/>}$
              $\xml{<equivalentClasses>}$ $\xml{<LHS>Forest</LHS>}$ $\xml{<RHS>ForestlandUse and MCON</RHS>}$ $\xml{</equivalentClasses>}$
            $\xml{</Box>}$
            $\xml{<Box>}$ $\xml{<Standpoint name=}\straightquote\xml{*}\straightquote\xml{/>}$
              $\xml{<subClassOf>}$ $\xml{<LHS>ForestlandUse</LHS>}$ $\xml{<RHS>Land</RHS>}$ $\xml{</subClassOf>}$
            $\xml{</Box>}$
          $\xml{</AND>}$ $\xml{</booleanCombination>}$ $\xml{</standpointLabel>}$
    \end{lstlisting}\end{small}
    \vspace{-2ex}
\end{example}

\myparagraph{Sharpening statements}
Sharpening statements \mbox{$\ste_1\preceq\ste_2$} are encoded via annotation of the ontology with a \standpointLabel \ of the form
\mbox{\(
    \xml{<Sharpening>} \mathrm{SPExpr} \ \mathrm{SPExpr} \xml{</Sharpening>}
\)}.

\myparagraph{Simple standpoint axioms}
Standard subclass and equivalence axioms can be turned into standpoint axioms by adding a \standpointLabel \ annotation of the form
\begin{align*}
    \mathrm{SPAxiom}        \mycoloneq& \      \xml{<standpointAxiom>} \mathrm{SPOperator} \xml{</standpointAxiom>} \ | \\
                            \phantom{=}& \     \xml{<standpointAxiom name=}\straightquote\xml{§ax}\straightquote\xml{>} \mathrm{SPOperator} \xml{</standpointAxiom>} \\
    \mathrm{SPOperator}     \mycoloneq& \      \xml{<Box>} \mathrm{SPExpr} \xml{</Box>} \ | \
                                               \xml{<Diamond>} \mathrm{SPExpr} \xml{</Diamond>}
\end{align*}
with $\xml{§ax}$ and $\mathrm{SPExpr}$ defined as above. This effectively prepends a standard subclass or equivalence axiom by a standpoint operator $\Box_{\mathsf{e}}$/$\Diamond_{\mathsf{e}}$ for some standpoint expression $\mathsf{e}$. If the name attribute of the \verb|standpointAxiom| element is given, it can be used as a reference in Boolean combinations. Note that a standpoint axiom with a name attribute will not be translated outside of the Boolean combinations that refer to them, since this would render any reference to it equivalent to ``true".


\section{Tool Description}\label{sec:tool-description}
    
Our command-line tool implements an adaptation to $\SROIQ$ of the translation from sentential Stand\-point-$\SROIQbs$ to standard $\SROIQbs$ proposed by G\'omez \'Alvarez et al.~\cite{alvarez:2022}. In this section, we describe how a standpoint-annotated OWL~2~DL ontology is translated to an OWL~2~DL ontology that can be processed by an OWL~2~DL reasoner. Subsequently, we explain how to use the command-line tool and outline some of its additional features.
\subsection{Implementation}\label{section:Implementation}
    
Our command-line tool\footnote{The source code can be found on the GitHub repository: \url{https://github.com/cl-tud/standpoint-owl2}} can parse a sentential Standpoint-$\SROIQ$ ontology in the syntax provided in Section~\ref{section:StandpointOWL2}, and translate it to standard OWL~2~DL, for which efficient reasoners already exist, e.g.\ HermiT~\cite{glimm:2014}. We use the \emph{OWL~API} 
for creating, parsing and manipulating OWL~2 ontologies, hence the format of the input ontology can be one of a variety of standardised syntaxes, such as RDF/XML~\cite{owl2RDFXML:2014} or Manchester syntax~\cite{horridge:2012}.

The implemented translation exploits the fact that satisfiable $\SSROIQ$ knowledge bases are guaranteed to have a model with a bounded number of precisifications, which are represented by integers $\pi \in \set{0,\ldots,p-1}$ in the encoding. While \citeauthor{alvarez:2022} \cite{alvarez:2022} set this bound to the size of the knowledge base, for our implementation we use the more fine-grained count of the diamonds occurring in positive polarity and the boxes occurring in negative polarity in the \standpointLabel \ annotations. 
There are two syntactic impediments of $\SROIQ$ that need to be addressed by the translation: (a) $\SROIQ$ does not provide nullary predicates, which we simulate by concept expressions of the form $\forall u.P$, where $u$ is the universal role and $P$ encodes the predicate via a concept name, and (b) $\SROIQ$ does not directly allow for arbitrary Boolean combinations of axioms, but an equivalent encoding is possible using the universal role $u$; for instance the expression $\neg(A \equiv B) \vee (A \sqsubseteq C)$ can be converted to $\top \sqsubseteq \forall u.(A \sqcap \neg B) \sqcup \forall u.(B \sqcap \neg A) \sqcup \forall u.(\neg A \sqcup C)$.

The translation proceeds in the following way. For each concept name $A$, role name $r$ and standpoint $s$ in the input ontology, we generate the fresh concept and role names $M\_s\_\pi$, $A\_\pi$ and $r\_\pi$ for each $\pi \in \{0,\ldots,p-1\}$, where $M$ is a prefix for the nullary standpoint predicates. To avoid altering the original ontology file, we additionally \emph{rebase} all concept, role and individual names, i.e.\ update their IRIs with that of the output ontology. 

Then, for each  $\pi \in \{0,\ldots,p-1\}$, we add the axioms 
${(
\top \sqsubseteq \forall u.M\_\xml{*}\_\pi
)}$
and for each standpoint axiom $\phi\in\mathcal{B}_T$ the set of GCIs consisting of
${(
\top \sqsubseteq \mathrm{trans}(\pi,\phi)
)}$, with $\mathrm{trans}$ defined as follows.
{\smaller
\begin{align*}
    \mathrm{trans}(\pi, C \sqsubseteq D)                 = & \ \forall u.(\neg C\_\pi \sqcup D\_\pi), &
    \mathrm{trans}_{\mathcal{E}}(\pi, s)                                   = & \ \forall u.M\_s\_\pi, \\
    \mathrm{trans}(\pi, \neg(C \sqsubseteq D))           = & \ \exists u.(C\_\pi \sqcap \neg D\_\pi), &
    \mathrm{trans}_{\mathcal{E}}(\pi, \mathsf{e}_1 \cap \mathsf{e}_2)      = & \ \mathrm{trans}_{\mathcal{E}}(\pi, \mathsf{e}_1) \sqcap \mathrm{trans}_{\mathcal{E}}(\pi, \mathsf{e}_2), \\
    \mathrm{trans}(\pi, \phi_1 \wedge \phi_2)            = & \ \mathrm{trans}(\pi, \phi_1) \sqcap \mathrm{trans}(\pi, \phi_2), &
    \mathrm{trans}_{\mathcal{E}}(\pi, \mathsf{e}_1 \cup \mathsf{e}_2)      = & \ \mathrm{trans}_{\mathcal{E}}(\pi, \mathsf{e}_1) \sqcup \mathrm{trans}_{\mathcal{E}}(\pi, \mathsf{e}_2), \\
    \mathrm{trans}(\pi, \phi_1 \vee \phi_2)              = & \ \mathrm{trans}(\pi, \phi_1) \sqcup \mathrm{trans}(\pi, \phi_2), &
    \mathrm{trans}_{\mathcal{E}}(\pi, \mathsf{e}_1 \setminus \mathsf{e}_2) = & \ \mathrm{trans}_{\mathcal{E}}(\pi, \mathsf{e}_1) \sqcap \neg\mathrm{trans}_{\mathcal{E}}(\pi, \mathsf{e}_2) \\
    \mathrm{trans}(\pi, \Box\nolimits_{\mathsf{e}})      = & \ \sqcap_{\pi' = 0}^{p-1} (\neg\mathrm{trans}(\pi',\mathsf{e}) \sqcup \mathrm{trans}(\pi',\phi)), \\
    \mathrm{trans}(\pi, \Diamond\nolimits_{\mathsf{e}})  = & \ \sqcup_{\pi' = 0}^{p-1} (\mathrm{trans}(\pi',\mathsf{e}) \sqcap \mathrm{trans}(\pi',\phi)),
\end{align*}}
Equivalence axioms are treated as a conjunction of subclass axioms, and negation in front of standpoint modalities is resolved by duality, viz.\
\mbox{\(
\neg\Box\nolimits_{\mathsf{e}}[\phi] = \Diamond\nolimits_{\mathsf{e}}[\neg\phi]
\)}
and
\mbox{\(
    \neg\Diamond\nolimits_{\mathsf{e}}[\phi] = \Box\nolimits_{\mathsf{e}}[\neg\phi]
\)}.
In line with treating plain $\SROIQ$ axioms as being prepended by $\Box_{*}$, we translate standard subclass and equivalence axioms as being of the form $\Box_{*}[\phi]$, and RIAs are translated by simply replacing the original role names $r$ by $r\_\pi$ for all precisifications $\pi \in \{0,\ldots,p-1\}$.
\subsection{Usage}\label{section:Usage}
    

\paragraph{Translate}
An annotated ontology can be directly translated via the command-line tool by providing the ontology file or its IRI.
When one or more of the options listed below are used, the output ontology will not be translated automatically, but saved in a separate file. This can be avoided by setting a separate \emph{translate} flag.

\myparagraph{Import}
The \emph{import} option first imports an ontology into the input file, and then annotates all imported axioms for which standpoint annotation is supported by a box operator with a specified standpoint name.
This feature avoids that, for instance, two concepts with the same name (and possibly different IRI bases) occuring in subclass or equivalence axioms will be treated as the same concept during translation.

\myparagraph{Query}
The most basic functionality of OWL~2~DL reasoners is checking the ontology for inconsistency.
Popular reasoners, e.g.\ HermiT~\cite{glimm:2014}, additionally offer to answer queries regarding subclass relations, instances etc.
However, these query services are impractical for translated Standpoint-OWL~2 ontologies, since standpoint axioms can only be used in a query if they are translated to standard OWL~2 beforehand.
In order to simplify the specification of queries containing standpoint axioms, we have added a \emph{query} option to the command-line tool. The query language is the language of Boolean combinations, i.e.\ we can ask whether a given Boolean combination is entailed by the translated ontology. A query can be given by an expression of the form $\mathrm{Formula}$ defined in Section~\ref{section:StandpointOWL2} (possibly in a separate file), or in a simplified query syntax for single standpoint axioms.
The syntax of a simple query is defined as follows:
\begin{align*}
    \mathrm{SimpleQuery}    \mycoloneq& \       \xml{[s]}(\mathrm{SimpleAxiom}) \ | \
                                                \xml{<s>}(\mathrm{SimpleAxiom}) \\
    \mathrm{SimpleAxiom}    \mycoloneq& \       \mathrm{Class} \ \xml{sub} \ \mathrm{Class} \ | \
                                                \mathrm{Class} \ \xml{eq}  \ \mathrm{Class}
\end{align*}
where $\xml{s}$ and $\mathrm{Class}$ are as before. The operators $\xml{[s]}$ and $\xml{<s>}$ stand for $\Box_{\mathsf{s}}$ and $\Diamond_{\mathsf{s}}$, respectively, $\xml{sub}$ for a subsumption relation and $\xml{eq}$ for equivalence of classes.
The query is first negated, and then added to the input ontology as a Boolean combination.
If, after translation, the resulting ontology is inconsistent, the query is a logical consequence of the ontology.

\myparagraph{Dump}
Lastly, there is the option to \emph{dump} the output ontology to the command-line, rather than saving it to a new file, which facilitates reasoning over the translated ontology via pipeline to an OWL~2~DL reasoner.

\section{Conclusion}
    
In this paper, we have proposed a syntax for sentential Stand\-point-$\SROIQ$ using OWL~2 annotations, which have proved useful for implementing different non-standard description logics. While in this paper we have focused on the sentential fragment of $\SROIQ$, our approach can be easily extended to more expressive fragments of standpoint $\SROIQ$, e.g.\ to support standpoint operators on the level of concepts and roles, which leads to fragments currently under investigation and with interesting applications in ontology alignment.
Subsequently, we have provided a translation from sentential Stand\-point-$\SROIQ$ to standard $\SROIQ$, which is an adjustment of the recently published translation for the more expressive $\SROIQbs$, and finally, we have implemented this translation as a command-line tool, thus effectively providing standpoint-based reasoning support for OWL~2 DL ontologies.

Future work will focus on the usability of the system. The XML syntax for \standpointLabel \ annotations is not user-friendly, and annotating an ontology in this way can be time-consuming, even when using an ontology editor like Prot\'eg\'e~\cite{musen:2015}. A possible approach to alleviating this problem would be to develop a plugin for Prot\'eg\'e and to make use of its existing user interface for adding and modifying standpoint axioms, similar to the Fuzzy~OWL~2 Prot\'eg\'e plugin by Bobillo and Straccia~\cite{bobillo:2011}. This will allow for the integration of the modelling support with the translator and reasoner.



\bibliography{sample-ceur}


\end{document}